\documentclass{article}
\usepackage{iclr2027_conference,times}

\usepackage{amsmath,amsfonts,bm}

\def\eqref#1{equation~\ref{#1}}

\def\1{\bm{1}}

\DeclareMathAlphabet{\mathsfit}{\encodingdefault}{\sfdefault}{m}{sl}
\SetMathAlphabet{\mathsfit}{bold}{\encodingdefault}{\sfdefault}{bx}{n}

\usepackage{hyperref}
\usepackage{url}
\usepackage{graphicx}
\usepackage{booktabs}   
\usepackage{multirow}
\usepackage{amsmath}
\usepackage{pifont}   

\usepackage{CJKutf8}
\newcommand{\cn}[1]{\begin{CJK*}{UTF8}{gbsn}#1\end{CJK*}}

\usepackage{xspace}
\newcommand{\method}{\textsc{TextVerifier}\xspace}   
\newcommand{\bench}{\textsc{TextRenderBench-ZH}\xspace}    

\title{\textsc{TextVerifier}: Seeing Beyond Recognition to Verify Chinese Text Rendering}
\author{Yiyang Chen\textsuperscript{1,2}\thanks{Work done during the internship at Alibaba Cloud Computing.}, 
    Guo-Hua Wang\textsuperscript{2}\thanks{Corresponding authors.}, 
    Yaoli Liu\textsuperscript{3,2}\footnotemark[1], 
    Qing-Guo Chen\textsuperscript{2}, \\
    \bfseries Zhao Xu\textsuperscript{2}, 
    Weihua Luo\textsuperscript{2}, 
    Xiujun Ma\textsuperscript{1}\footnotemark[2]\\[4pt]
    \normalfont\textsuperscript{1} State Key Laboratory of General Artificial Intelligence, Peking University, Beijing, China\\
    \normalfont\textsuperscript{2} Alibaba Group\\
    \normalfont\textsuperscript{3} State Key Lab of CAD\&CG, Zhejiang University\\  
    \normalfont\texttt{chenyy@stu.pku.edu.cn, wangguohua@alibaba-inc.com, maxiujun@pku.edu.cn}
}
\iclrfinalcopy   

\begin{document}
\maketitle

\begin{abstract}
Text-to-image models increasingly generate readable Chinese text that still contains local
rendering defects, such as a character with an extra or missing stroke or a malformed component.
Such errors are invisible to OCR and vision--language models, which are trained for recognition
and are therefore \emph{invariant} to the very structural deviations that define a rendering error.
We formalize \emph{text rendering verification} as a target-conditioned problem: given a target
character and an image crop, decide whether the crop is a correct rendering of that character. Our
verifier \method\ feeds the text tower a character's Ideographic Description Sequence (IDS) rather
than a single-character symbol, and is trained on $12.7$M crop--IDS pairs plus label-free
``same-character, different-quality'' hard pairs sampled from the training trajectory of a
text-region inpainting model. To evaluate this ability we build \bench, the first per-character
benchmark for generated Chinese text, annotating position, identity, and correctness for $7{,}358$
characters. \method\ reaches an F1 of $0.844$, far above OCR confidence ($0.526$), a general VLM
($0.329$), and TextPecker ($0.590$). Used as a per-character reward for Flow-GRPO post-training, it
further reduces rendering errors, showing that its score can both detect and correct errors.
\end{abstract}

\section{Introduction}
\label{sec:intro}

\begin{figure}[t!]
\centering
\includegraphics[width=0.85\linewidth]{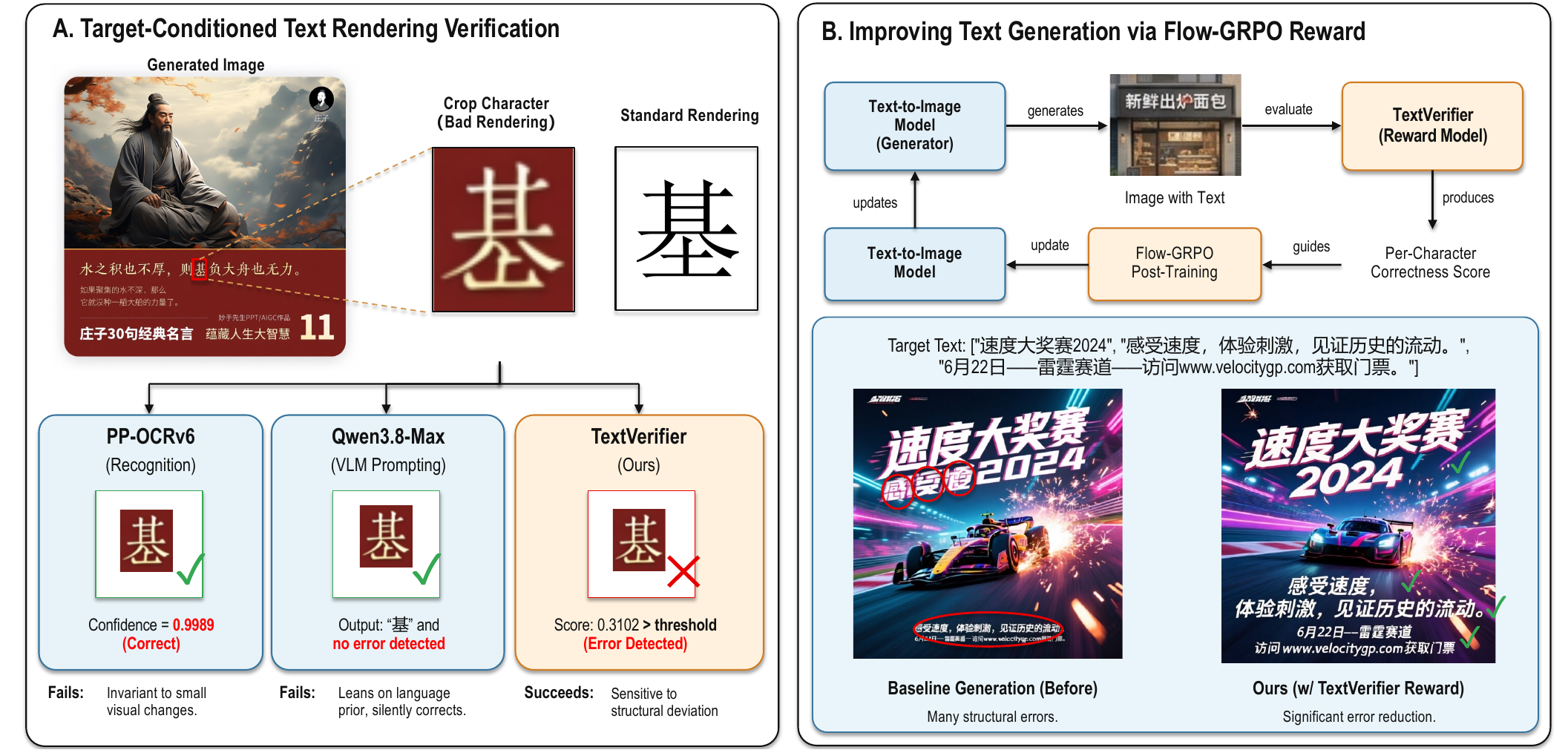}
\caption{\textbf{Text rendering verification and its use as a reward.} (A)~For the character ``\cn{基}'' in a generated image, OCR and VLM both miss the error, whereas \method\ correctly flags it
as an error. (B)~Using the per-character score as a post-training reward markedly reduces glyph errors.}
\vspace{-17pt}
\label{fig:teaser}
\end{figure}

Rendering text inside images has become a core capability of text-to-image models.
Recent foundation models \citep{wu2025qwenimage,bytedanceseed2026seedream5,openai2026gptimage25,googledeepmind2026gemini31flashliteimage} render full
sentences and even paragraphs of English and Chinese, and text rendering is now a key
axis along which their quality is compared \citep{li2026qwenimagebench,chang2025oneigbench,geng2025xomni}. Even so,
the strongest models still make rendering errors, and whether an error can be caught
depends on the script: in English a wrong glyph changes the recognized word, so an OCR
model flags it, whereas a malformed Chinese character is usually still read as the
intended one and slips through. This asymmetry matters because Chinese, with six to
eight thousand densely stroked characters built from components under hierarchical
spatial structures \citep{zhang2018radical,cao2020zeroshot}, fails not as an unreadable string but as a readable
sentence with a single flawed character: a headline reads correctly at a glance, yet one
character adds or drops a stroke, or malforms a component (Figure~\ref{fig:teaser}). In
publishing, education, and other zero-tolerance settings, one such mistake
can make an entire image unusable. Evaluating or correcting such a model therefore rests
on a more basic ability: to automatically judge whether every
Chinese character in a generated image is rendered correctly.

The most natural idea is to reuse a text recognizer, yet recognition and
correctness verification pursue opposite objectives. OCR models are trained for content
recovery: they read which character is present despite changes in font, style,
deformation, and blur, and is thus deliberately invariant to a glyph's visual
variation. Verification demands the opposite: sensitivity to structural deviation, so
that a glyph with an added or missing stroke, or a blurred component, stays
distinguishable from the canonical form. PP-OCRv6 \citep{zhang2026ppocrv6} illustrates the mismatch: it
assigns confidence above $0.99$ even to a clearly wrong character (Figure~\ref{fig:teaser}),
so its confidence is a poor criterion for rendering correctness. Prompting a general
vision-language model directly is no better: on our benchmark even the best-performing
Gemini~3.7~Flash \citep{googledeepmind2026gemini37flash} recalls only $54\%$ of the erroneous characters. Recognition models,
in short, do not provide a reliable criterion for glyph correctness, which instead calls for a
capability trained specifically for structural verification.


TextPecker \citep{zhu2026textpecker} is the first work to address this problem in a
systematic way, fine-tuning an 8B vision-language model as a structural-anomaly
assessor. However, it folds blur, deformation, misalignment, and
glyph error into a single notion of structural anomaly and judges an image without
conditioning on the intended character. Lacking any model of which renderings of a
character are legitimate, it readily flags correct characters as
anomalies: on our benchmark it reaches a recall of $0.765$ at a precision of only
$0.480$ (F1 $0.590$), producing a very high number of false alarms. Its 8B
backbone also makes it costly to run at scale or to use as a training-time reward.

We therefore formalize the task as \emph{target-conditioned text rendering verification}.
Given a target character $c$ supplied by a recognizer (e.g., an OCR model) and the corresponding image crop $x$,
the goal is to decide whether $x$ is a correct rendering of $c$. This is naturally
a conditional anomaly-detection problem: each character spans a manifold of
legitimate renderings across fonts, calligraphy, and artistic styles, and an
incorrect glyph lies a small distance off that manifold. Guided by this notion, we build \method, a CLIP-style dual-tower model
\citep{radford2021learning} whose text tower takes the Ideographic Description Sequence (IDS)
\citep{unicode2026standard} of a character rather than a single-character token. We assemble an
automatic labeling pipeline that yields $12.7$M crop--IDS pairs from about $2.5$M real images
for the first-stage alignment pretraining. For discriminative supervision, we construct a set of
hard positive--negative pairs by two routes: manual synthesis, and mining
``same-character, different-quality'' pairs from the training trajectory of a text-region
inpainting model. We then train the verifier to produce a continuous, calibrated,
target-conditioned score for each character.

To evaluate this ability we build \bench, the first per-character benchmark for
Chinese text rendering, labeling every character with its location, identity, and correctness.
On it \method\ reaches an F1 of $0.844$, well above OCR, general-VLM, and TextPecker \citep{zhu2026textpecker} baselines.
Used as a per-character reward for Flow-GRPO \citep{liu2025flowgrpo} post-training, it further
reduces rendering errors in a text-to-image model, showing that the score can not only detect
errors but also help remove them.


We summarize our contributions as follows: 


\begin{itemize}
  \item We build \method, a structure-aware verifier that encodes each character's IDS
  decomposition in a CLIP-style dual tower, reaching high accuracy at only $0.36$B
  parameters.
  \item We design a three-stage training curriculum and a label-free way to construct its
  supervision, combining synthetic positive--negative pairs with hard pairs mined from
  a text-region inpainting model's training trajectory.
  \item We build \bench, the first per-character benchmark for generated Chinese text verification,
  on which \method\ reaches an F1 of $0.844$ at $7.6$~ms per character, and show that it
  can serve as a per-character reward for Flow-GRPO training.
\end{itemize}


\section{Related Work}
\label{sec:related}

\paragraph{Text recognition.}
Text recognition aims to recover \emph{which} characters an image contains, spanning
two-stage detect-then-recognize pipelines \citep{baek2019character,zhang2026ppocrv6}, end-to-end
vision-language parsers \citep{wei2024generalocr,wei2025deepseekocr,zhang2026paddleocrvl16,lu2026ovisocr2}, and
general-purpose VLMs \citep{qwen2026qwen38max}. Whatever the architecture, these models
optimize recognition accuracy and are thus invariant to font, style, and degradation:
a glyph with an added stroke is read as the same character with high
confidence, and VLM-based systems further lean on a language prior that quietly
corrects ill-formed inputs \citep{liang2026visualmerit,tong2024eyes}. We
therefore separate recognition from verification: OCR supplies only the target
character and its location, and correctness is left to a dedicated verifier
trained for sensitivity rather than invariance.

\paragraph{Chinese structural representation.}

Chinese characters decompose into many components under a small set of
spatial structures, written linearly as an Ideographic Description Sequence (IDS)
\citep{unicode2026standard}. This compositionality drives zero-shot recognition through
radical or stroke analysis \citep{zhang2018radical,cao2020zeroshot} and, closest to us, CLIP-style
image--IDS alignment for a canonical per-character representation \citep{yu2023chinese}.
A related line checks whether a handwritten character is well formed, decoding an image
into an IDS and flagging any mismatch with a known character \citep{li2023tree,hu2025count,sun2024multimodal},
recently extended to full text lines \citep{li2024towards}. These target recognition or
handwriting, where glyphs are clean and errors are whole-component swaps. Our setting
differs on three counts. Generation errors are pixel-level rather than component-level:
broken or adhered strokes and stylized fonts leave a glyph just off the manifold of valid
renderings rather than turning it into a different character. The target character is
already supplied by an upstream recognizer, so we verify a known identity instead of
predicting it. And our supervision is label-free rather than hand-annotated.

\paragraph{Benchmarks and rewards for text rendering.} Visual text rendering has progressed from glyph-conditioned control
\citep{yang2023glyphcontrol,tuo2024anytext} to native rendering inside foundation models
\citep{wu2025qwenimage,bytedanceseed2026seedream5}, yet its evaluation still rests almost entirely on OCR
string metrics over prompt-only benchmarks \citep{chen2023textdiffuser,du2025textcrafter}, which
misread stylized-but-correct text, are sensitive to segmentation, and are judged by
VLMs that hallucinate correctness \citep{hu2025amo,liang2026wegenbench}. Models of glyph quality are
recent: TextPecker fine-tunes an 8B VLM for structural-anomaly detection
\citep{zhu2026textpecker}, TIQA regresses perceptual quality while ignoring correctness
\citep{koltsov2026tiqa}, and Type-R detects typos by word-level OCR differencing \citep{shimoda2025typer};
none labels each generated character with position, identity, and correctness \citep{li2024towards}.
The same string-versus-glyph gap reappears in reward design for post-training
\citep{black2024training,wallace2024diffusiondpo,liu2025flowgrpo}: OCR-string rewards are cheap but blind to glyph
form and easily gamed \citep{liu2025flowgrpo,qiao2026firered,wang2026gdro}, whereas VLM-judge rewards spend a
large-model forward pass per sample and return discrete verdicts
\citep{zhu2026textpecker,cui2026textalign}, and another line uses
human-annotated region-level preferences \citep{shuai2026glyphprinter}. Our verifier instead is
target-conditioned, per-character, continuous, and calibrated, and complements rather
than replaces the OCR semantic reward inside Flow-GRPO.

\begin{figure}[t]
\centering
\includegraphics[width=0.95\linewidth]{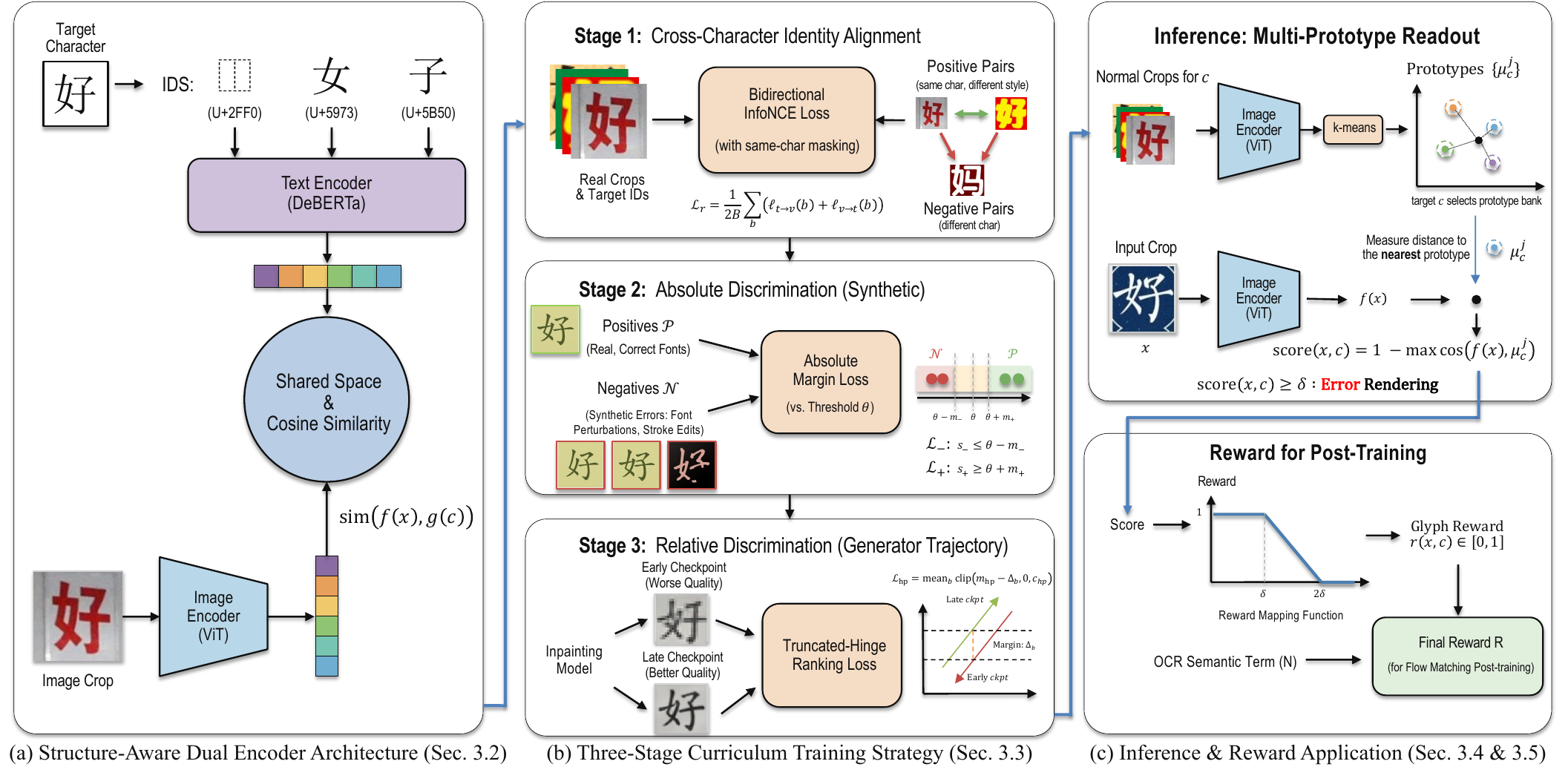}
\caption{\textbf{Structure-aware dual tower and three-stage training.} (a)~The structure-aware CLIP-style dual encoder model architecture. (b)~The three-stage training curriculum.
(c)~Inference procedure, and it can also serve as a reward for Flow-GRPO post-training.}
\label{fig:arch}
\vspace{-16pt}
\end{figure}

\section{Method}
\label{sec:method}

\subsection{Problem Formulation and Overview}
Given a target character $c$ and its image crop $x$, the verifier outputs a
continuous score $\mathrm{score}(x,c)$ and decides through a threshold $\delta$
whether $x$ is a correct rendering of $c$; here $c$ is a condition of the decision
rather than a prediction target, and we only judge whether $x$ belongs to the
legitimate writings of $c$. The problem is thus a per-character conditional anomaly
detection: across fonts, calligraphy, and styles a character spans a family of
legitimate renderings (its ``correct'' manifold), and an error is only a small
departure from it. 

\subsection{Structure-Aware Dual Encoders}
\label{sec:arch}
We base the verifier on a CLIP-style dual tower \citep{radford2021learning} but reshape its text
side. Since the ``text'' here is only the target character itself, treating each
character as an independent token would reduce the text tower to a class lookup
table with no structural relation between characters. We instead feed it the
character's Ideographic Description Sequence (IDS) \citep{unicode2026standard}: with a dozen
structural operators and about a thousand components, an IDS writes each character
uniquely as a sequence describing its components and structure
(Figure~\ref{fig:arch}), so that characters sharing components or structure lie
close at the sequence level and structural differences can be encoded explicitly.
The text tower adopts the CCR-CLIP \citep{yu2023chinese} decomposition table with a
from-scratch DeBERTa \citep{he2021deberta}; the image tower is a ViT \citep{dosovitskiy2021vit} whose
patch tokens are averaged into the image feature. Each tower has an MLP projection
head into a shared, $\ell_2$-normalized space, and the cross-modal match is the
cosine $\mathrm{sim}(f(x),g(c))$.

\subsection{Three-Stage Data Construction}
\label{sec:data}
\begin{figure}[t]
\centering
\includegraphics[width=1.00\linewidth]{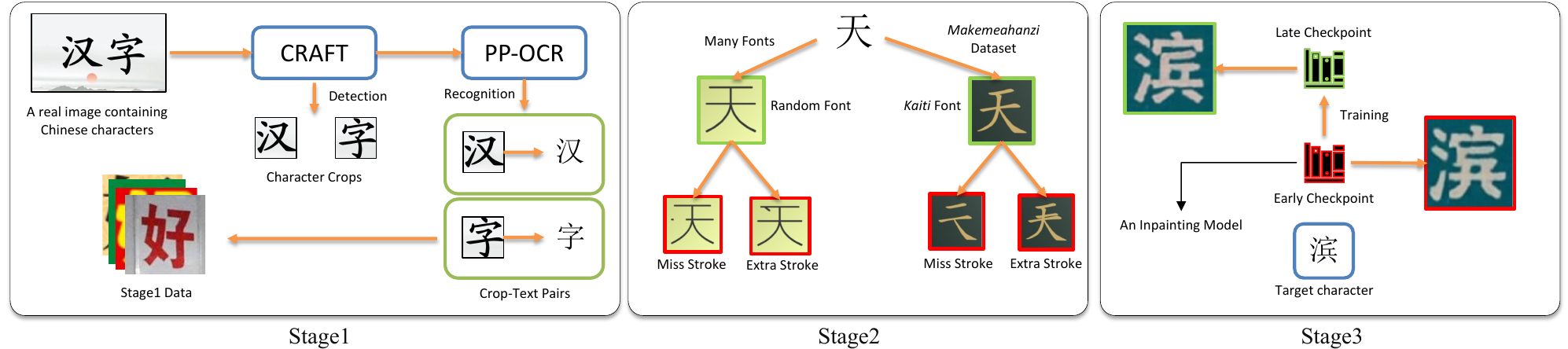}
\caption{\textbf{Three-stage data construction.} \emph{Left:} real images pass through a
data pipeline to yield stage1 data. \emph{Middle:}
standard glyphs are perturbed by two stroke edits into pairs. \emph{Right:} early and late checkpoints of a text-region inpainting
model repaint the same target character.}
\label{fig:data_pipeline}
\vspace{-6pt}
\end{figure}

The verifier is trained in three stages over three kinds of data. The first stage teaches the
model to distinguish the identities of different characters, using other characters in the same
batch as negatives. With in-batch negatives, however, the model separates clearly
different characters but cannot notice the fine stroke-level errors within a
single character. The second and third stages therefore both introduce \emph{hard negatives} for
the same character, i.e., wrong renderings that closely resemble the correct glyph and differ only
in small stroke or structural details; the two stages mainly differ in how these hard negatives are
constructed, while also using different label semantics and loss forms. Figure~\ref{fig:data_pipeline}
overviews the three-stage data construction.

The first stage draws data from real images containing Chinese text, via an automatic pipeline of
two off-the-shelf models: CRAFT \citep{baek2019character} is a character-level detector that
locates each character and gives its bounding box, and PP-OCRv6 \citep{zhang2026ppocrv6} then
recognizes the crop in each box to obtain its target character. From about $2.5$M real images this
yields over ten million crop--character pairs covering $8{,}189$ characters. As character frequency
is heavily long-tailed, we pad rare characters with font-rendered standard glyphs to at least $500$
samples each, giving about $12.73$M crop--IDS pairs. Details are shown in Appendix~\ref{A.2}.

The second stage builds hard negatives by applying controlled perturbations to standard glyphs
along two complementary routes. Font synthesis renders each character's standard glyph in about
thirty fonts and applies coarse perturbations: a missing stroke, which erases a segment of stroke
pixels and fills it with background, or an added stroke, which overlays a horizontal, vertical,
left-slanting, or right-slanting mark. The Makemeahanzi \citep{kishore2016makemeahanzi} dataset
provides per-stroke vector paths for the kaiti font, supporting structurally precise
whole-stroke edits such as deleting a full stroke or grafting one from another character into light
contact with the ink. The two routes produce about $1.75$M
glyph images; examples are shown in Appendix~\ref{A.3}.

The third stage builds negatives that more closely resemble real generation errors, which can only
come from a generator itself. We exploit a convenient fact: a text-region inpainting model still in
training leaves, along its trajectory, ``same-character, different-quality'' samples. For the same
location and target character, its early checkpoints often render strokes wrongly while its late
checkpoints are mostly correct, with texture, background, and color largely unchanged and only the
glyph differing. We train such an inpainting model and use one early and one late checkpoint to
repaint the same target character, yielding about $0.7$M pairs. The inpainting model's architecture
and further sample pairs are given in Appendix~\ref{A.4}.

\subsection{Three-Stage Training}
\label{sec:curriculum}
The three stages use the three kinds of data from Sec.~\ref{sec:data}, with the loss form set by
the nature of each stage's labels.

\paragraph{Stage 1: cross-character identity alignment.} Using the crop--IDS pairs, we align the
image embedding $v_b=f(x_b)$ with the character IDS embedding $t_b=g(c_b)$ by a
bidirectional InfoNCE loss:
\begin{equation}
\ell_{t\to v}(b)=-\log\frac{\exp(\mathrm{sim}(t_b,v_b)/\tau)}
{\sum_{j:\,c_j\neq c_b\,\lor\,j=b}\exp(\mathrm{sim}(t_b,v_j)/\tau)},\quad
\mathcal{L}_{r}=\tfrac{1}{2B}\sum_{b}\big(\ell_{t\to v}(b)+\ell_{v\to t}(b)\big).
\end{equation}
where $\ell_{v\to t}$ is symmetric and $\tau$ is a temperature. Here the negatives are the other
characters in the same batch. The only change from standard CLIP is \emph{same-character masking}:
other crops of the same character in a batch are excluded from the negatives, since they are
further legitimate points on that character's manifold. This stage learns identity only and cannot
yet tell apart fine stroke errors of the same character.

\paragraph{Stage 2: absolute discrimination from synthetic samples.} This stage uses the synthetic
perturbation samples to introduce hard negatives for the same character. Anchored on the target
character, the positive set $\mathcal{P}$ contains font-rendered correct glyphs, real crops, and the
kaiti glyph, and the negative set $\mathcal{N}$ contains the synthesized wrong glyphs. These
labels are \emph{absolute}: positives are unconditionally correct and negatives are necessarily
wrong. We therefore use a threshold $\theta$ to anchor same-character positives and negatives to its
two sides, each with a margin $m_+$ or $m_-$:
\begin{equation}
\mathcal{L}_{+}=\underset{b,\,f\in\mathcal{P}_b}{\mathrm{mean}}\ \mathrm{softplus}
\!\Big(\tfrac{\theta+m_{+}-\mathrm{sim}(t_b,f)}{\tau_\theta}\Big),\quad
\mathcal{L}_{-}=\underset{b,\,f\in\mathcal{N}_b}{\mathrm{mean}}\ \mathrm{softplus}
\!\Big(\tfrac{\mathrm{sim}(t_b,f)-\theta+m_{-}}{\tau_\theta}\Big).
\end{equation}
and replay Stage 1's $\mathcal{L}_r$ to avoid forgetting identity. This stage begins to separate
correct from incorrect, but the synthetic negatives still differ in distribution from real errors,
so the learned boundary is loose, which leads to Stage 3.

\paragraph{Stage 3: relative discrimination from a generator's trajectory.} This stage uses the
early/late checkpoint pairs, taking the late render as the positive and the early one as the
negative. Unlike Stage 2, these labels are only \emph{relative}: we cannot assert that the late
render is correct or the early one wrong, only that the late is no worse than the early. We
therefore drop the absolute-threshold anchoring of Stage 2 and use only a truncated-hinge (ramp)
loss to keep the positive above the negative; with
$\Delta_b=\mathrm{sim}(t_b,f_b^{+})-\mathrm{sim}(t_b,f_b^{-})$,
\begin{equation}
\mathcal{L}_{\mathrm{hp}}=\underset{b}{\mathrm{mean}}\ \mathrm{clip}
\big(m_{\mathrm{hp}}-\Delta_b,\;0,\;c_{\mathrm{hp}}\big).
\end{equation}
Once a pair is correctly ordered ($\Delta_b\ge m_{\mathrm{hp}}$) the gradient is
zero, which avoids treating a relative signal as an absolute target; the cap $c_{\mathrm{hp}}$ suppresses noisy pairs with
very negative $\Delta$. Stage 3 adds $\mathcal{L}_{\mathrm{hp}}$ on top of all earlier
losses and converges in a single epoch.

\subsection{Inference: Multi-Prototype Readout}
\label{sec:infer}
At test time, thresholding $\mathrm{sim}(f(x),g(c))$ directly is unreliable:
per-character baselines vary, and a character's legitimate writings are multi-modal. We instead keep
several \emph{prototypes} per character in image space to describe its legitimate
manifold. For character $c$, its normal crops (sampled from the Stage-1 data) are
encoded and clustered by $k$-means \citep{lloyd1982least} into $K$ prototypes $\{\mu_c^{j}\}$, and the score
is the cosine distance to the nearest one:
\begin{equation}
\mathrm{score}(x,c)=1-\max_{j}\ \cos\!\big(f(x),\,\mu_c^{j}\big).
\end{equation}
A larger score means $x$ is farther from any normal writing of the character. The
threshold is a single quantile value $\delta$: we pool the scores of all normal
training crops and take a high quantile (e.g., the 90th percentile). The readout
costs only $K$ cosine similarities per character.

\subsection{The Verifier as a Reward for Post-training}
\label{sec:reward}
A continuous per-character score is a natural reward. Given a generated
image, the same front-end (CRAFT \citep{baek2019character} $+$ PP-OCRv6 \citep{zhang2026ppocrv6}) crops its text regions and aligns them
to the target characters of the prompt, and each is scored. We map a character's
score to a reward in $[0,1]$ that is $1$ for a clearly correct glyph, $0$ for a
clearly wrong one, and linear in between:
\begin{equation}
r(x,c)=\mathrm{clip}\!\Big(2-\tfrac{\mathrm{score}(x,c)}{\delta},\,0,\,1\Big)=
\begin{cases}
1, & \mathrm{score}\le\delta,\\
2-\mathrm{score}/\delta, & \delta<\mathrm{score}<2\delta,\\
0, & \mathrm{score}\ge 2\delta.
\end{cases}
\end{equation}
Over all target characters $\{c_j\}$ of an image, the glyph term weights the mean
with the worst case, the latter mildly penalizing an image that is acceptable overall
but badly wrong on a single character:
\begin{equation}
G=0.9\cdot\underset{j}{\mathrm{mean}}\,r(x_j,c_j)+0.1\cdot\underset{j}{\min}\,r(x_j,c_j).
\end{equation}
As a complement, a recognition-based semantic term $N=1-\mathrm{NED}\in[0,1]$ (the
normalized edit distance between the OCR reading and the target text) constrains
\emph{what} is written. The final glyph-dominated reward is
\begin{equation}
R=0.8\,G+0.2\,N,
\end{equation}
which plugs into a flow-matching post-trainer such as Flow-GRPO \citep{liu2025flowgrpo}.

\section{Experiments}
\label{sec:exp}

\subsection{\bench}

\begin{figure}[t]
\centering
\includegraphics[width=0.95\linewidth]{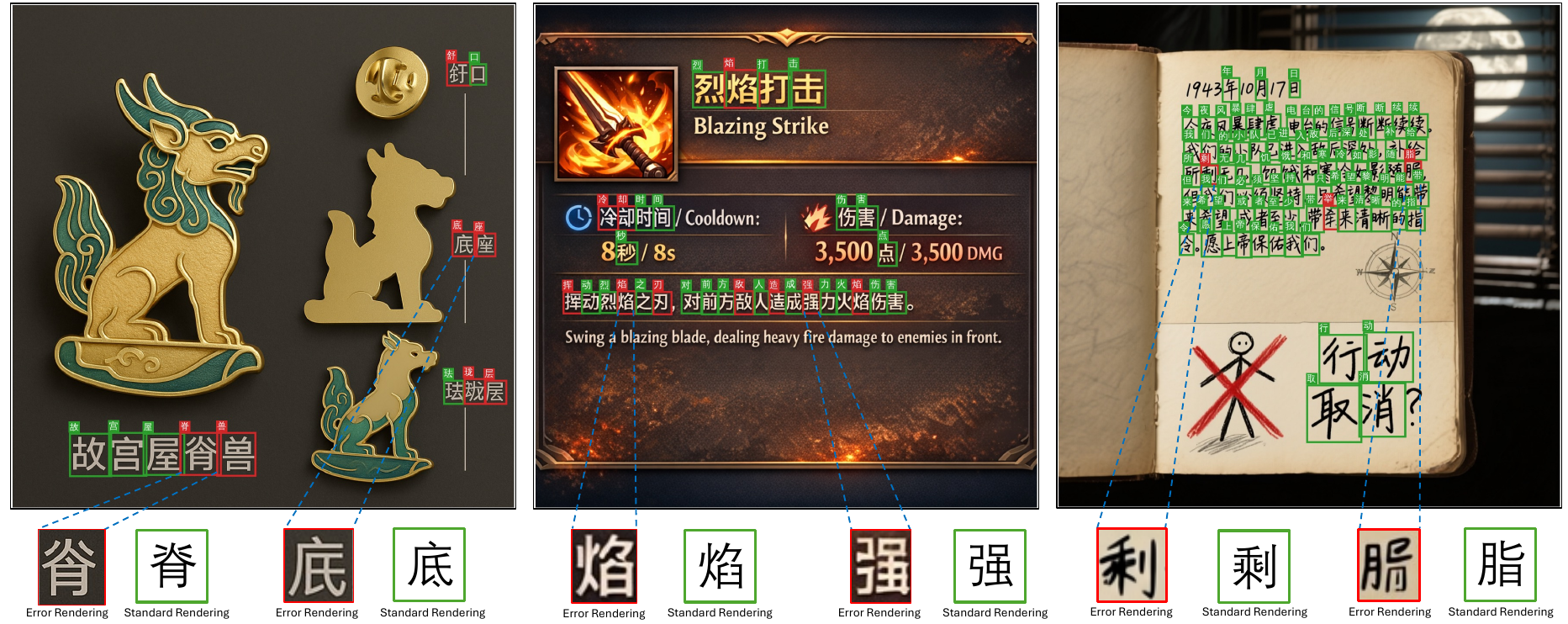}
\caption{\textbf{Per-character annotation examples from \bench.} Images are generated by
text-to-image models; each character carries a (position, identity,
correctness) triple (green box: correct, red box: error), with error-versus-standard glyph
pairs enlarged below.}
\label{fig:bench}
\vspace{-16pt}
\end{figure}

Existing text-rendering benchmarks provide only prompts and score whole images
with OCR string metrics; they can neither judge glyph correctness character by
character nor evaluate a verifier itself. We therefore build \bench, the first
per-character benchmark of Chinese text rendering. Its test images are generated
by text-to-image models: from $308$ images we collect $7{,}358$ character crops, $642$ of them confirmed by human
annotators as rendering errors. Each character carries a
(position, identity, correctness) triple (Figure~\ref{fig:bench}); Appendix~\ref{app:qual} details the
two-step annotation protocol and the error criterion. This is the first
benchmark for generated images with per-Chinese-character triple annotation; it measures a
verifier's character-level precision and recall. 

\subsection{Setup}
The image tower is a ViT-L/16 \citep{dosovitskiy2021vit} , and the text tower is a
DeBERTa-v2 large \citep{he2021deberta} over the $981$-symbol IDS vocabulary; both project to a $1024$-dimensional shared
space. Each stage continues from the previous stage's weights: Stages 1 and 2 train
for two epochs each and Stage 3 for one epoch. The loss hyperparameters are: InfoNCE
temperature $\tau{=}0.3$; Stage-2 threshold loss $\theta{=}0.7$, $m_+{=}0.1$,
$m_-{=}0.2$, $\tau_\theta{=}0.05$; Stage-3 ramp $m_{\mathrm{hp}}{=}0.05$,
$c_{\mathrm{hp}}{=}0.2$. For each character we cluster $500$ normal crops into
$K{=}10$ prototypes and score a crop by its cosine distance to the nearest
prototype; the global threshold is the $90$th percentile of the training
distribution, giving $\delta{=}0.0134/0.0188/0.0214$ for the three stages.

A rendering error is the positive class and a
correct glyph the negative in our benchmark, so a false positive (FP) is a false alarm on a correct
glyph and a false negative (FN) a missed error.
Baselines are PP-OCRv6 \citep{zhang2026ppocrv6} character confidence (a glyph is flagged when its confidence
falls below a threshold; we report an oracle best threshold as an upper reference),
the general VLMs Qwen3.8-Max \citep{qwen2026qwen38max}, GPT-5.6-sol \citep{openai2026gpt56}, and Gemini~3.7~Flash \citep{googledeepmind2026gemini37flash}, and the two released
TextPecker \citep{zhu2026textpecker} assessors.

\subsection{Main Results}
Table~\ref{tab:main} reports all methods, and every baseline is far from usable: PP-OCRv6 \citep{zhang2026ppocrv6}
confidence peaks at only $0.53$ F1 even at an oracle threshold; the general VLMs either almost
never fire (Qwen3.8-Max \citep{qwen2026qwen38max}, GPT-5.6-sol \citep{openai2026gpt56}) or fire often but wrongly (Gemini~3.7~Flash \citep{googledeepmind2026gemini37flash}); and
TextPecker \citep{zhu2026textpecker} raises a very high number of false alarms (F1 $0.590$, InternVL3-8B variant
$0.433$). \method\ reaches F1 $0.844$ after all three
stages (precision $0.830$, recall $0.858$), far above every baseline. Beyond accuracy, the
verifier is compact and fast (last two columns of Table~\ref{tab:main}): its scoring model is $0.36$B
parameters, $23\times$ smaller than TextPecker \citep{zhu2026textpecker} and far below the closed frontier VLMs,
yet it is the most accurate.
At $7.6$~ms per character (median, full test set, batch$=$1; Appendix~\ref{app:impl} details the
measurement setup), it is also $2.3\times$ faster than PP-OCRv6 \citep{zhang2026ppocrv6} recognition ($17.2$~ms) and orders
of magnitude faster than TextPecker \citep{zhu2026textpecker} VLMs.

\begin{table}[t]
\caption{\textbf{Main results on \bench}. \method's three-stage F1 of $0.844$ is far above all baselines;
the ``best thr'' row tunes the PP-OCRv6 confidence \citep{zhang2026ppocrv6} threshold on the test set as an upper reference only.
Params is the scoring-model size; Inf. is per-character median latency (ms) on the full test set.}
\label{tab:main}
\centering
\resizebox{\linewidth}{!}{%
\begin{tabular}{lrrrrrrrrr}
\toprule
Method & TP↑ & FP↓ & FN↓ & TN↑ & P↑ & R↑ & F1↑ & Params & Inf.(ms)↓ \\
\midrule
PP-OCRv6 (thr $=0.90$) \citep{zhang2026ppocrv6}       & 223 & 140 & 419 & 6576 & 0.614 & 0.347 & 0.444 & \multirow{2}{*}{0.06B} & \multirow{2}{*}{17.2} \\
PP-OCRv6 (best thr) \citep{zhang2026ppocrv6}     & 400 & 479 & 242 & 6237 & 0.455 & 0.623 & 0.526 & & \\
\midrule
Qwen3.8-Max \citep{qwen2026qwen38max}                   &  78 & 66 & 564 & 6650 & 0.542 & 0.121 & 0.198 & --- & --- \\
GPT-5.6-sol \citep{openai2026gpt56}                         & 37 & 57 & 605 & 6659 & 0.394 & 0.058 & 0.101 & --- & --- \\
Gemini 3.7 Flash \citep{googledeepmind2026gemini37flash}                     & 347 & 1122 & 295 & 5594 & 0.236 & 0.540 & 0.329 & --- & --- \\
\midrule
TextPecker (Qwen3-VL-8B) \citep{zhu2026textpecker}              & 491 & 531 & 151 & 6185 & 0.480 & 0.765 & 0.590 & 8.29B &  320.4 \\
TextPecker (InternVL3-8B) \citep{zhu2026textpecker}            & 322 & 525 & 320 & 6191 & 0.380 & 0.502 & 0.433 & 8.1B &  276.0 \\
\midrule
Ours: Stage 1                         & 477 &  \textbf{71} & 165 & \textbf{6645} & \textbf{0.870} & 0.743 & 0.802 & \multirow{3}{*}{0.36B} & \multirow{3}{*}{\textbf{7.6}} \\
Ours: Stage 2                         & 484 &  76 & 158 & 6640 & 0.864 & 0.754 & 0.805 & & \\
\textbf{Ours: Stage 3}                & \textbf{551} & 113 & \textbf{91} & 6603 & 0.830 & \textbf{0.858} & \textbf{0.844} & & \\
\bottomrule
\end{tabular}%
}
\end{table}

\subsection{Ablations}

\paragraph{IDS structural prior.} Replacing the text-tower input from IDS with a
single-character token (one token per character, everything else fixed) and
retraining all three stages (Table~\ref{tab:ids}) drops F1 by about $10.4$, $7.1$,
and $10.1$ points across the stages, with the loss falling almost entirely on recall
(Stage-3 recall from $0.858$ to $0.648$) while precision even rises slightly: without
a structural prior the model sees only ``clearly different'' samples and loses much of its sensitivity to
stroke-level deviations. Appendix~\ref{app:ids} further shows that the IDS tower's embedding geometry is aligned
with structural similarity and that its recall advantage holds across all
character-frequency strata.

\paragraph{Training stages and loss terms.} In Stage 3 three losses coexist: the new
ramp pairwise loss, the Stage-2 threshold loss, and the Stage-1 InfoNCE \emph{replayed}
to preserve cross-character identity. Table~\ref{tab:stages} removes them one at a time.
Dropping the Stage-1 InfoNCE replay costs $4.8$ points ($0.844\!\to\!0.796$), dropping
the Stage-2 threshold loss $2.7$ points ($\to\!0.817$), and skipping Stage 2 entirely
(Stage 1 straight to Stage 3) $3.3$ points ($\to\!0.811$); with no Stage-1 pretraining
(random initialization) the model collapses to $0.111$, as discriminative training
relies on a representation already organized by glyph shape. Each term contributes to performance. Every ablation keeps recall near $0.86$ but lowers precision
markedly, confirming Sec.~\ref{sec:method}: the ramp pairwise loss alone constrains
neither the representation's overall scale nor its cross-character separability, which
are exactly what the replay and threshold losses supply.

\begin{table}[t]
\centering
\begin{minipage}[t]{0.53\linewidth}
\centering
\caption{\textbf{Text-tower input: IDS vs.\ a single-character token.} A single-character token
lowers F1 by $\sim$$10.4/7.1/10.1$ across the three stages.}
\label{tab:ids}
\small\setlength{\tabcolsep}{4pt}
\begin{tabular}{llrrr}
\toprule
Input & Stage & P & R & F1 \\
\midrule
IDS          & 1 & \textbf{0.870} & 0.743 & 0.802 \\
single-char  & 1 & 0.862 & 0.586 & 0.698 \\
IDS          & 2 & 0.864 & 0.754 & 0.805 \\
single-char  & 2 & 0.879 & 0.631 & 0.734 \\
\textbf{IDS} & \textbf{3} & 0.830 & \textbf{0.858} & \textbf{0.844} \\
single-char  & 3 & 0.870 & 0.648 & 0.743 \\
\bottomrule
\end{tabular}
\end{minipage}\hfill
\begin{minipage}[t]{0.44\linewidth}
\centering
\caption{\textbf{Training stages and loss terms}. \textbf{Full} vs.\ removing one
component: \ding{192} skip Stage 2; \ding{193} Stage 3 w/o Stage2 threshold loss; \ding{194} Stage 3
w/o Stage-1 loss replay; \ding{195} no Stage-1 pretraining.}
\label{tab:stages}
\small\setlength{\tabcolsep}{4pt}
\begin{tabular}{crrr}
\toprule
Config & P & R & F1 \\
\midrule
\textbf{Full} & \textbf{0.830} & \textbf{0.858} & \textbf{0.844} \\
\ding{192} & 0.767 & 0.861 & 0.811 \\
\ding{193} & 0.774 & 0.864 & 0.817 \\
\ding{194} & 0.746 & 0.852 & 0.796 \\
\ding{195} & 0.080 & 0.181 & 0.111 \\
\bottomrule
\end{tabular}
\end{minipage}
\end{table}

\begin{figure}[t]
\centering
\includegraphics[width=0.95\linewidth]{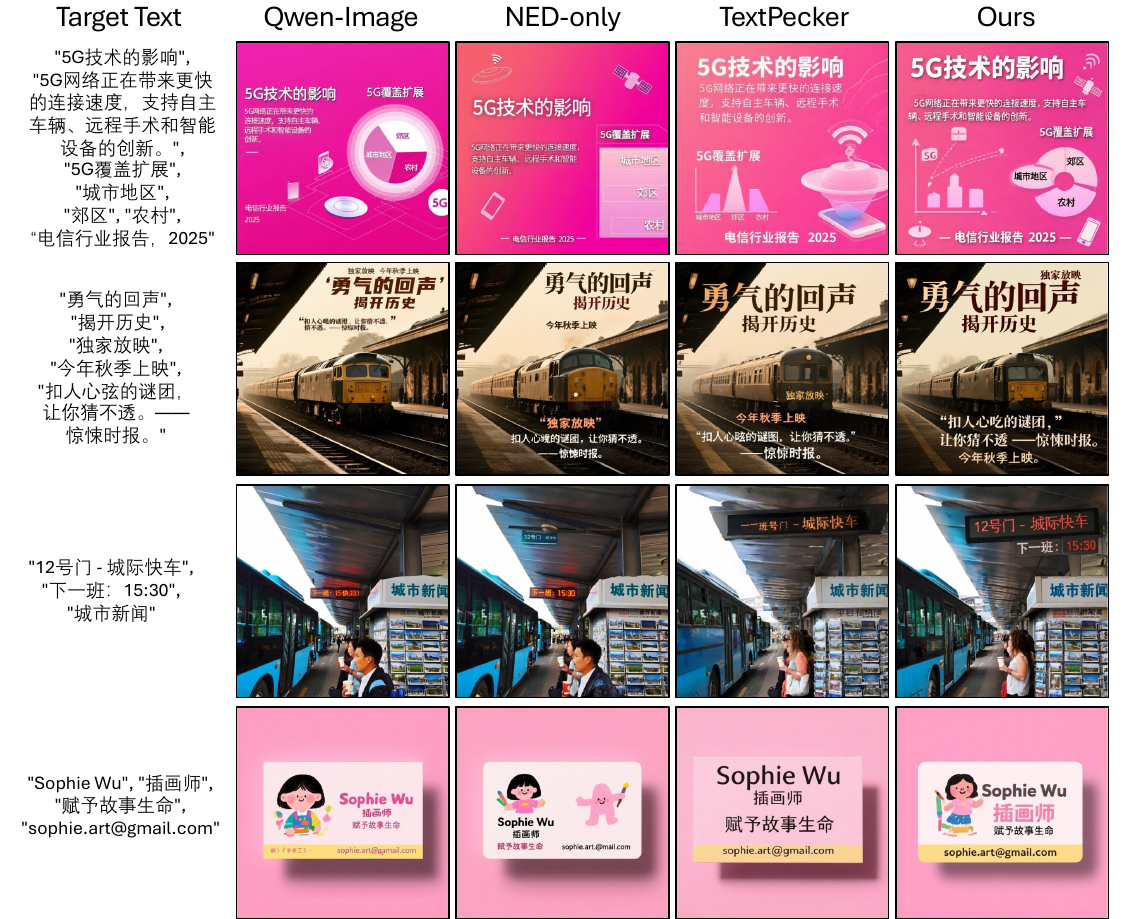}
\caption{\textbf{Qualitative comparison of Flow-GRPO post-training under different rewards.}
Given the target text, we compare Qwen-Image \citep{wu2025qwenimage}, the NED-only ablation,
TextPecker \citep{zhu2026textpecker}, and ours; Ours yields fewer glyph errors while
staying aesthetically closer to Base.}
\label{fig:grpo}
\end{figure}

\subsection{Reward for Flow-GRPO}
\label{sec:rewardexp}
We further use the verifier as a post-training reward. Starting from Qwen-Image \citep{wu2025qwenimage}, we post-train with Flow-GRPO \citep{liu2025flowgrpo}, feeding
$R$ to its reward interface($\beta{=}10^{-3}$, $20$
denoising steps for rollout and training, otherwise default), and evaluate on oneID-ZH \citep{chang2025oneigbench}.
RL prompts follow the Chinese text-rendering corpus of \citet{zhu2026textpecker}, restricted to
its harder, text-dense subset ($\geq 35$ characters) to speed up training.

Our aim is to reduce glyph errors while disturbing the base generator as little as possible.
Table~\ref{tab:reward} reports glyph accuracy (Glyph-Acc, the fraction of target characters our
\method judges correctly rendered), diversity (DreamSim \citep{fu2023dreamsim}), aesthetics
(ImageReward \citep{xu2023imagereward}), and the reward model's cost; Base (Qwen-Image \citep{wu2025qwenimage}) is the
un-post-trained reference, which trivially keeps the highest diversity and aesthetics but also all
of its rendering errors. Two findings emerge. First, fixing glyphs requires a glyph-aware reward:
a recognition-only reward (NED-only, dropping the glyph term $G$) barely lifts Glyph-Acc over Base
($0.812\!\to\!0.822$, Table~\ref{tab:reward_abl}), whereas the full reward reaches $0.915$. Second,
against TextPecker \citep{zhu2026textpecker}, our reward attains comparable
glyph accuracy ($0.915$ vs.\ $0.919$) while perturbing diversity and aesthetics less ($0.175$ vs.\
$0.165$; $0.633$ vs.\ $0.629$), with a $23\times$ smaller reward model ($0.36$B vs.\ $8.29$B).
Among methods that actually reduce
rendering errors, then, ours is the least disruptive and the most efficient.
Figure~\ref{fig:grpo} shows qualitative comparisons.

\begin{table}[t]
\centering
\begin{minipage}[t]{0.635\linewidth}
\centering
\caption{\textbf{Flow-GRPO post-training under different rewards} (oneID-ZH; $\uparrow$ higher
is better). Base (Qwen-Image \citep{wu2025qwenimage}) is the un-post-trained reference (not bold); between the two
rewards that fix glyphs, the better is bold.}
\label{tab:reward}
\small\setlength{\tabcolsep}{4pt}
\resizebox{\linewidth}{!}{%
\begin{tabular}{lcccc}
\toprule
Reward & Glyph-Acc$\uparrow$ & DreamSim$\uparrow$ & ImageReward$\uparrow$ & Reward Params$\downarrow$ \\
\midrule
Base & 0.812 & 0.2023 & 0.6645 & -- \\
\midrule
TextPecker & \textbf{0.919} & 0.1651 & 0.6288 & 8.29B \\
\textbf{Ours} & 0.915 & \textbf{0.1753} & \textbf{0.6330} & \textbf{0.36B} \\
\bottomrule
\end{tabular}%
}
\end{minipage}\hfill
\begin{minipage}[t]{0.34\linewidth}
\centering
\caption{\textbf{Reward ablation.} A recognition-only reward (i.e., drop the glyph
term $G$) barely improves Glyph-Acc over Base; the glyph term drives the gain.}
\label{tab:reward_abl}
\small\setlength{\tabcolsep}{4pt}
\begin{tabular}{lc}
\toprule

Reward & Glyph-Acc$\uparrow$ \\
\midrule

NED-only (drop $G$) & 0.822 \\
\textbf{Ours} (full $R$) & \textbf{0.915} \\
\bottomrule
\end{tabular}
\end{minipage}
\end{table}

\section{Conclusion}
\label{sec:conclusion}
We formalized Chinese text-rendering verification in generated images as target-conditioned
anomaly detection, and built \method, an IDS-conditioned dual tower trained with a three-stage
curriculum of cross-character alignment, synthetic absolute anchoring, and relative discrimination
from an inpainting model's trajectory. On \bench, it reaches $0.844$ F1 at $7.6$~ms per character. As a
Flow-GRPO reward, it matches 8B-VLM rewards on error reduction while better preserving diversity
and aesthetics.

\subsection*{Ai Use Statement}
We used generative AI tools only for language polishing of author-written text.
All technical content, experiments, and claims were produced and verified by the
authors, who take full responsibility for the final content.


\bibliography{references}
\bibliographystyle{iclr2027_conference}

\newpage
\appendix

\section{Implementation Details}
\label{app:impl}
This section collects configuration and data details omitted from the main text for
space; all stages continue from the previous stage's weights.

\subsection{Architecture and Training Hyperparameters}
Table~\ref{tab:impl} summarizes the configuration. Loss hyperparameters are restated
from Sec.~\ref{sec:exp} for a single reference point.

\begin{table}[h]
\centering
\caption{Training configuration.}
\label{tab:impl}
\small
\begin{tabular}{p{0.26\linewidth}p{0.66\linewidth}}
\toprule
Component & Setting \\
\midrule
Image tower & ViT-L/16, input $128\times128$, patch $16$ ($64$ patch tokens, mean-pooled); $306.4$M params \\
Text tower & DeBERTa-v2 large, trained from scratch; IDS vocabulary $981$ ($12$ operators, $965$ components, $4$ special) \\
Projection & per-tower MLP head to a $1024$-d, $\ell_2$-normalized shared space \\
Epochs & Stage 1 / 2 / 3 $=$ $2$ / $2$ / $1$ \\
Batch per step & $96$ threshold units, $160$ replay samples, $128$ hard pairs \\
Loss hyperparams & $\tau{=}0.3$; $\theta{=}0.7$, $m_+{=}0.1$, $m_-{=}0.2$, $\tau_\theta{=}0.05$; $m_{\mathrm{hp}}{=}0.05$, $c_{\mathrm{hp}}{=}0.2$ \\
\bottomrule
\end{tabular}
\end{table}

\paragraph{Inference timing setup.} Inference latency reported in Table~\ref{tab:main} was measured
on the full $7{,}358$-character test set, batch$=$1, on $4\times$ NVIDIA H800 (80GB) GPUs. Ours
(Stage~3, bf16, ViT-L/16@128) and PP-OCRv6 \citep{zhang2026ppocrv6} (fp32, medium-rec with IR \& memory optimization) ran
on dedicated GPUs in parallel, each warmed up with 100 iterations before timing. Reported values
are per-character median milliseconds over the full test set, including all preprocessing, H2D
transfer, forward pass, and synchronization (``crop in $\to$ result out''), ensuring a fair
comparison. TextPecker's \citep{zhu2026textpecker} per-character cost is dominated by the 8B vision-encoder prefill per
image, and thus substantially higher than both Ours and PP-OCRv6 \citep{zhang2026ppocrv6}.

\subsection{Stage-1 Data Pipeline}
\label{A.2}
We run CRAFT \citep{baek2019character} detection and PP-OCRv6 \citep{zhang2026ppocrv6} recognition over about $2.5$M Chinese images and
filter crop--text pairs for reliability: OCR confidence $\ge 0.98$, $\min(w,h)\ge 48$px,
and aspect ratio $\le 2.0$. This yields $10.13$M crop--text pairs over $8{,}189$ distinct
characters (the $6{,}500$ level-1/2 characters plus out-of-list ones). The distribution is
heavily long-tailed (most frequent character $\approx$$110$K crops, least frequent $1$;
mean $1{,}237$, median $129$); rare characters are padded with font-rendered standard
glyphs so that every character has at least $500$ crops, giving about $12.73$M crop--IDS
pairs for the Stage-1 InfoNCE.

\subsection{Stage-2 Synthesis}
\label{A.3}
\begin{figure}[h]
\centering
\includegraphics[width=0.95\linewidth]{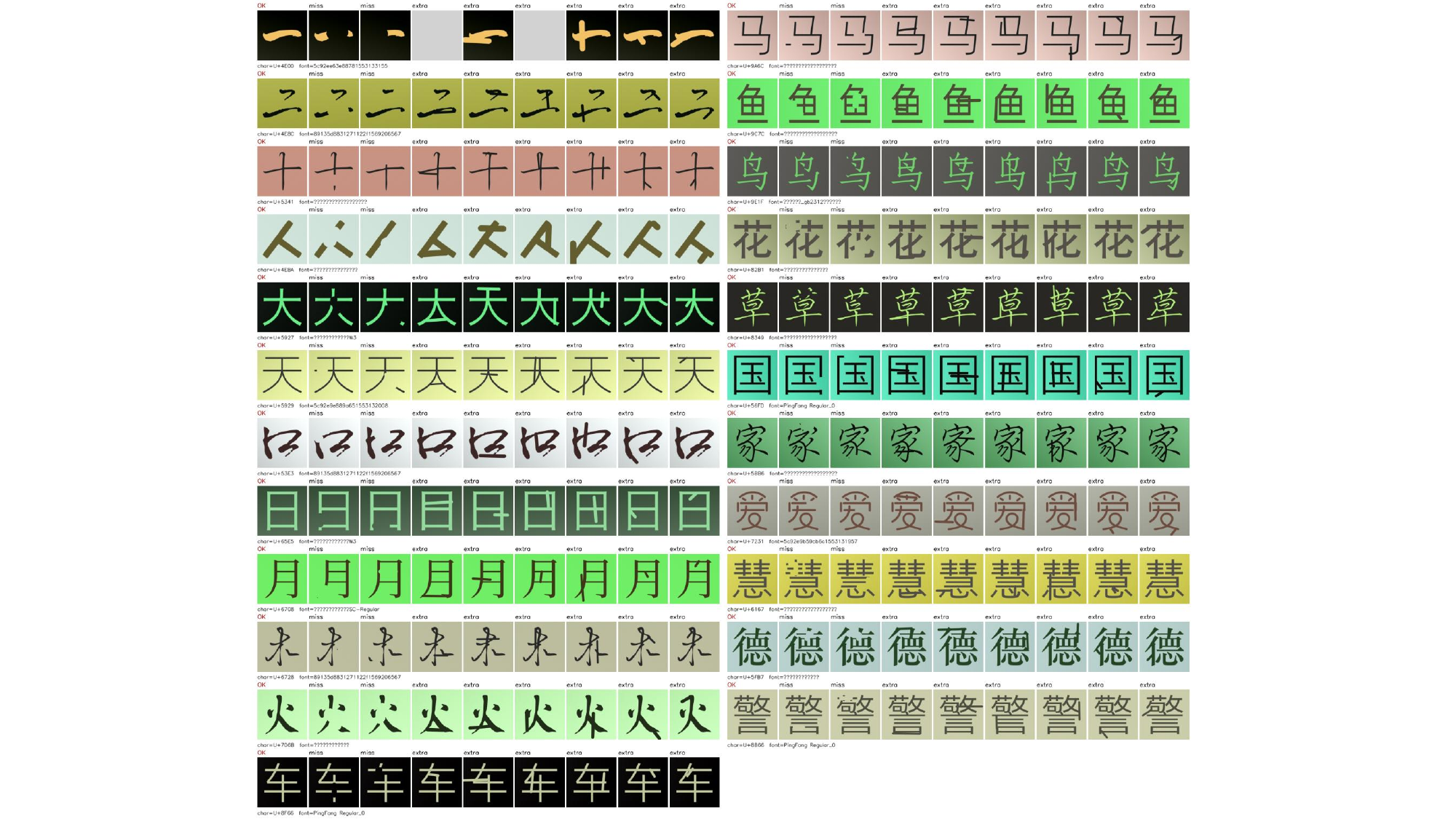}
\caption{The positive and negative samples for stage‑2 synthesized using synthetic font files. The first column in each row corresponds to the positive sample, which represents the correct standard rendering of the font. The remaining columns are negative samples containing perturbations such as stroke addition and stroke removal.}
\label{fig:stage2-1}
\end{figure}

\begin{figure}[h]
\centering
\includegraphics[width=0.95\linewidth]{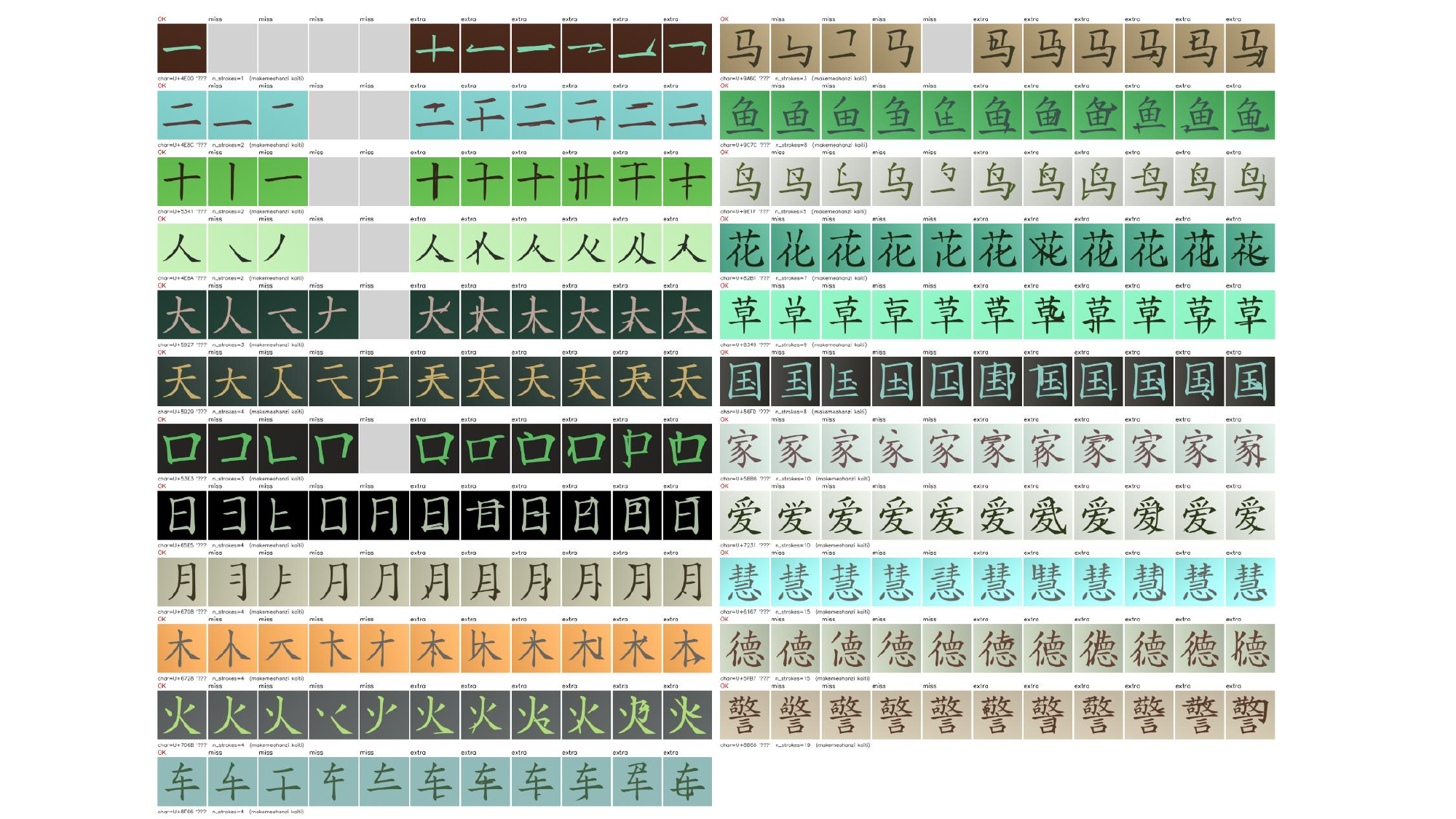}
\caption{The stage‑2 positive and negative samples synthesized using Chinese kaiti‑style vector data included in \emph{Makemeahanzi} \citep{kishore2016makemeahanzi}. The first column of each row is the positive sample, representing the correct standard rendering of the font, and the rest are negative samples with perturbations such as stroke addition and stroke removal.}
\label{fig:stage2-2}
\end{figure}

Stage-2 pairs a character's correct and incorrect renderings as absolute-label positives
and negatives; negatives are synthesized by two complementary routes, about $1.75$M images
in total. (i) \emph{Font synthesis}: as shown in figure~\ref{fig:stage2-1}, we render the standard glyph in $30$+ fonts, then apply
coarse perturbations---missing strokes (erase a stroke's pixels, fill with background) and
extra strokes (overlay one of a horizontal / vertical / left- / right-slanted stroke). 
(ii) \emph{Makemeahanzi synthesis}~\citep{kishore2016makemeahanzi}: this dataset gives per-stroke
vector decompositions of the standard (kaiti) font, enabling semantically precise whole-stroke
edits (delete a full stroke, or graft a stroke from another character and shift it into light
contact with the target ink), i.e.\ finer-grained negatives. We illustrate the construction of these positive and negative samples in Figure~\ref{fig:stage2-2}. Positives $\mathcal{P}$ are
font-correct renders, real crops, and the kaiti glyph; negatives $\mathcal{N}$ are the
synthesized errors. The Stage-1 InfoNCE $\mathcal{L}_r$ is replayed to preserve cross-character
identity.

\subsection{Stage-3 Inpainting Model}
\label{A.4}

\begin{figure}[h]
\centering
\includegraphics[width=0.95\linewidth]  {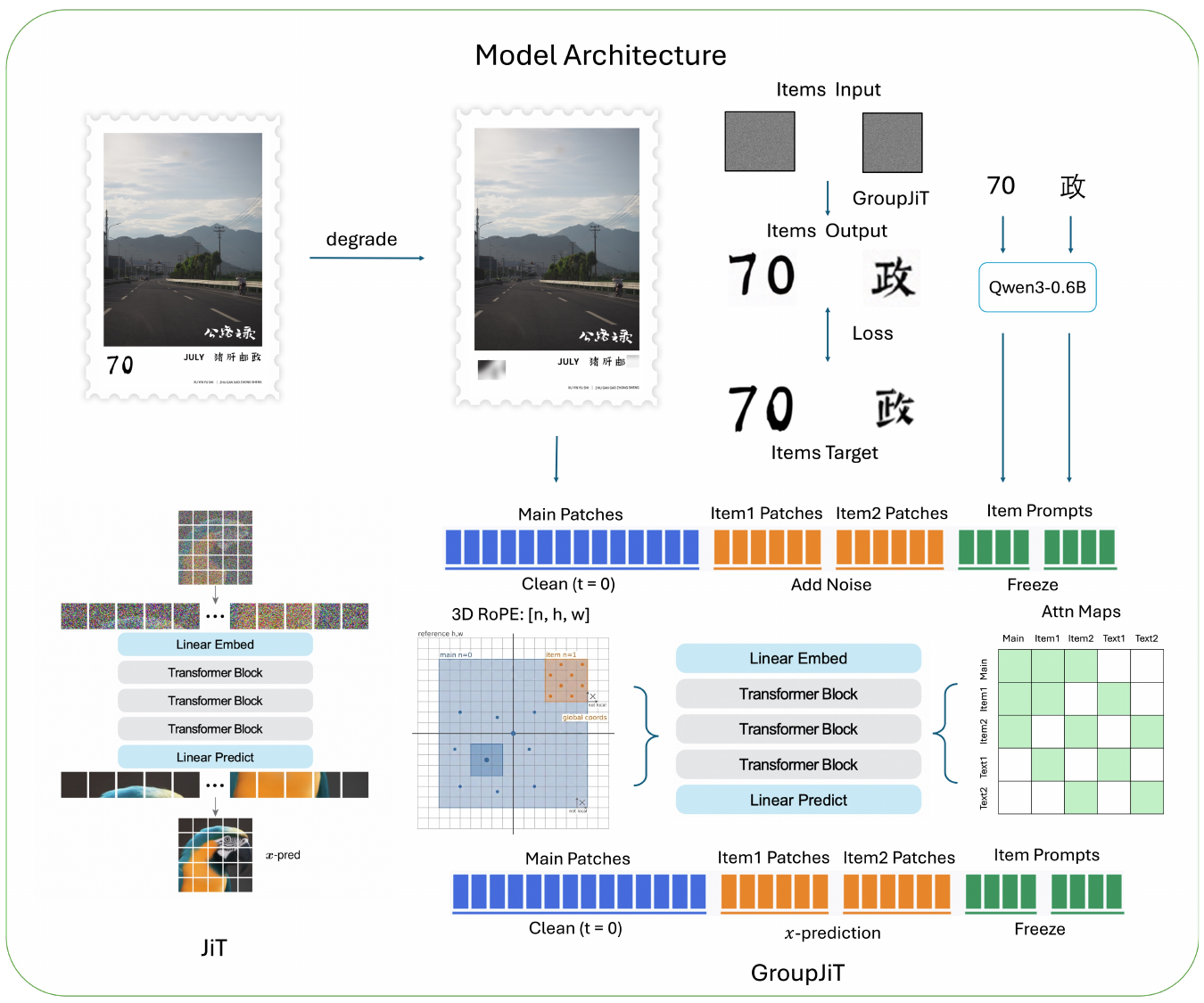}
\caption{Architecture of the text-region inpainting model for Stage-3 relative pair construction.
Three token groups, namely the full image, multiple target crops and character embeddings from Qwen3-0.6B \citep{yang2025qwen3}, are concatenated along the sequence dimension.
Custom attention mask and modified 3D-RoPE are adopted to enable multi‑crop training with absolute positional awareness.}
\label{fig:groupjit}
\end{figure}

The Stage‑3 relative pairs are derived from the training trajectory of a text‑region inpainting model. Built upon the JiT~\citep{li2026jit} architecture (pixel‑space, VAE‑free; JiT‑H/16, $\approx 1$B params), the model concatenates three token groups along the sequence dimension: the full image (the region to be repainted is corrupted with Gaussian noise while the rest remains clean, serving as conditioning), target crops (the flow‑matching target outputs, of which there can be multiple), and the target character's Qwen3‑0.6B~\citep{yang2025qwen3} embedding (frozen and used as conditioning). Its architectural design is illustrated in figure~\ref{fig:groupjit}. Following JiT \citep{li2026jit}, it adopts the x‑prediction MSE loss to compute losses over target crop regions for training the flow‑matching model.

Specifically, since the model is capable of structuring multiple target crops together with their corresponding character‑embedding prompts, we devise an attention mask as shown in figure~\ref{fig:groupjit}. This mask ensures that when computing self‑attention, each target crop can only attend to the full image, its own tokens, and its own embedding, and is prevented from attending to other crops. Furthermore, to inform the model of where each target crop is located within the full image, we modify 2D‑RoPE \citep{heo2024ropevit}. We first introduce an additional $n$ dimension to distinguish the full image ($n=0$) from target crops ($n=1$). Second, we map both the full image and all target crops onto the absolute coordinate system of a large shared canvas to obtain the $h$ and $w$ coordinates for each patch, instead of taking the top‑left corner of each individual image as the origin of the relative coordinate system as in vanilla 2D‑RoPE \citep{heo2024ropevit}. Accordingly, we supply the model with absolute position information for every patch of each target crop via 3D‑RoPE, so that the model can perceive their positions inside the full image.

\begin{figure}[h]
\centering
\includegraphics[width=\linewidth]{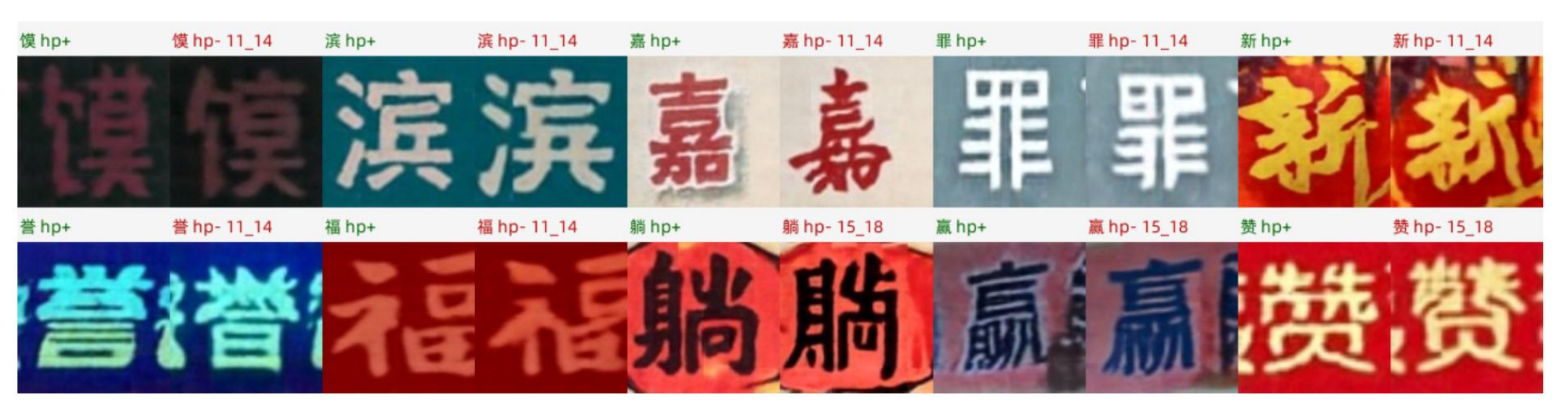}
\caption{The Stage‑3 positive and negative samples generated from different checkpoints of the inpainting model. In each pair, the first one is the positive sample produced by the later checkpoint, and the second one is the negative sample. For each paired positive‑negative sample, the positive sample exhibits better quality than its negative counterpart.}
\label{fig:stage3}
\end{figure}

Its training data can be fully reused from Stage~1 without additional annotation. We repaint the same (position, target‑character) triple using an early checkpoint and a late checkpoint, yielding around $0.7$M pairs, as shown in figure~\ref{fig:stage3}. Outputs from the late checkpoint are treated as positive samples and those from the early checkpoint as negative samples. Since the labels are purely relative (``the late‑checkpoint output is no worse than the early‑checkpoint one''), Stage~3 employs the truncated‑hinge loss $\mathcal{L}_{\mathrm{hp}}$ instead of a loss with absolute thresholds.

\subsection{Inference Readout and Threshold}
Comparing $\mathrm{sim}(f(x),g(c))$ directly is unreliable: per-character match baselines vary,
and a character's legal renderings are multi-modal, so collapsing them to one direction blurs the
gaps between modes. We instead keep $K$ image-space prototypes per character to trace its legal
manifold, scoring a crop by cosine distance to the nearest prototype; the threshold $\delta$ is a
high percentile ($90$th) of all normal-crop training scores and thus depends only on the training
distribution, not the test set. For reproducibility the k-means \citep{lloyd1982least} seed follows
\texttt{seed = K*100 + char\_index}.

\section{Structural Evidence for the IDS Text Tower}
\label{app:ids}
A label-free analysis of the trained IDS and single-character towers supports the claim
(Sec.~\ref{sec:exp}) that IDS yields a structured representation.

\begin{figure}[h]
\centering
\includegraphics[width=\linewidth]{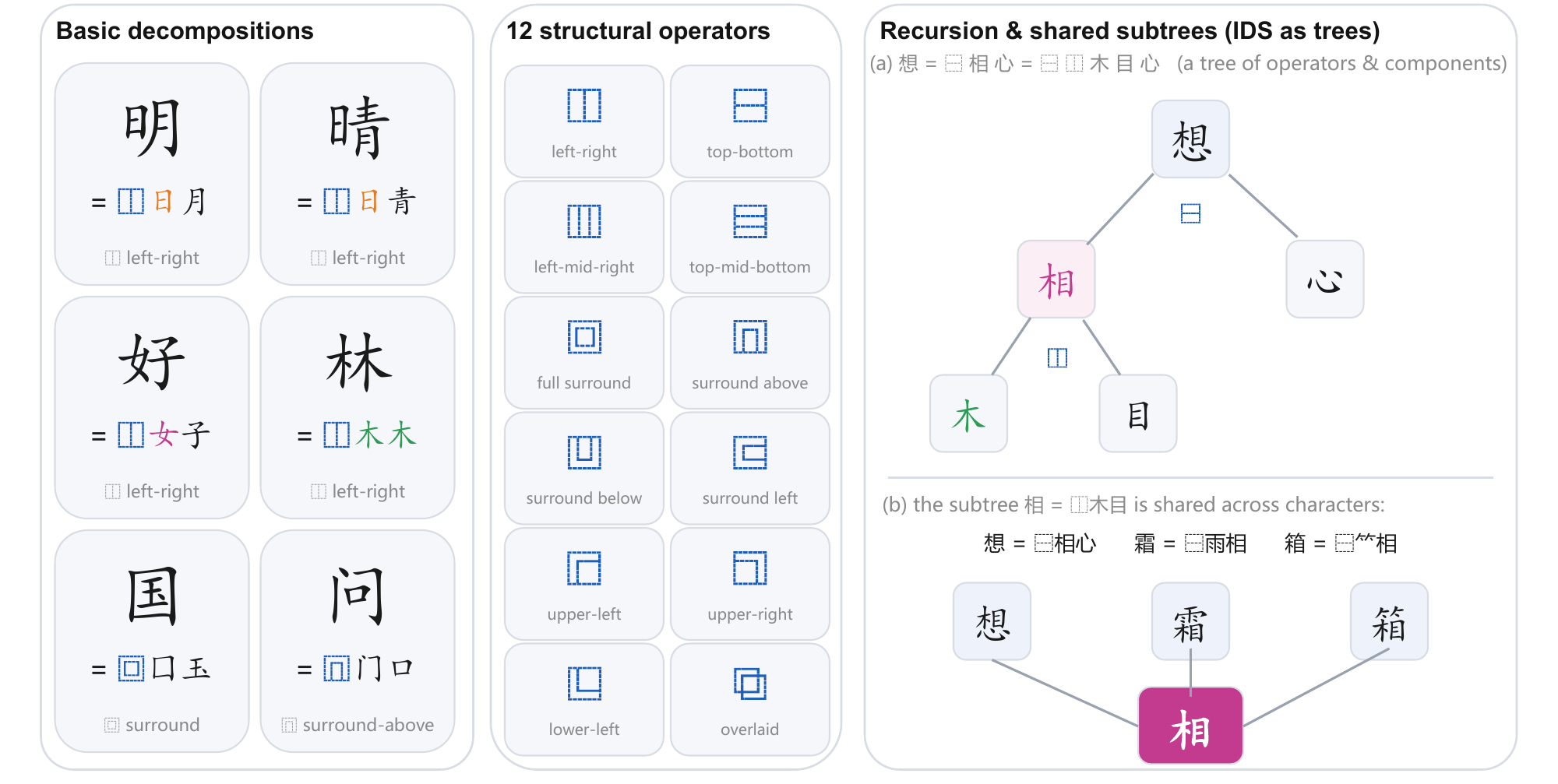}
\caption{\textbf{The IDS representation of Chinese characters.} \emph{Left:} basic
decompositions, where components shared across characters are color-coded (\cn{日}, \cn{女},
\cn{木}), so structurally related characters overlap at the symbol level. \emph{Middle:} the
twelve IDS structural operators. \emph{Right:} IDS is recursive---a character expands into a
tree of operators and components (\cn{想}$=$\cn{相}\cn{心}$=$\cn{木}\cn{目}\cn{心})---and
identical subtrees are reused across characters (\cn{相}$=$\cn{木}\cn{目} recurs in \cn{想},
\cn{霜}, \cn{箱}). Structural operators are shown only inside the figure.}
\label{fig:ids_repr}
\end{figure}

\paragraph{IDS decomposition (illustration).} Figure~\ref{fig:ids_repr} illustrates the IDS
representation from three angles. Basic decompositions write each character uniquely as an
operator applied to components (e.g.\ \cn{明}$=$\cn{日}\cn{月} under a left--right operator),
with components that recur across characters shown in a shared color, so that structurally
related characters visibly overlap at the symbol level. The twelve structural operators cover
the layouts a decomposition can take. Finally, an IDS is recursive: a character expands into a
tree whose subtrees are themselves valid decompositions (\cn{想}$=$\cn{相}\cn{心}$=$\cn{木}\cn{目}\cn{心}),
and identical subtrees are reused across characters (the subtree \cn{相}$=$\cn{木}\cn{目} recurs
in \cn{想}, \cn{霜}, and \cn{箱}). This shared, compositional structure is exactly what the IDS
text tower encodes and what a single-character token discards.

\begin{figure}[t]
\centering
\includegraphics[width=0.9\linewidth]{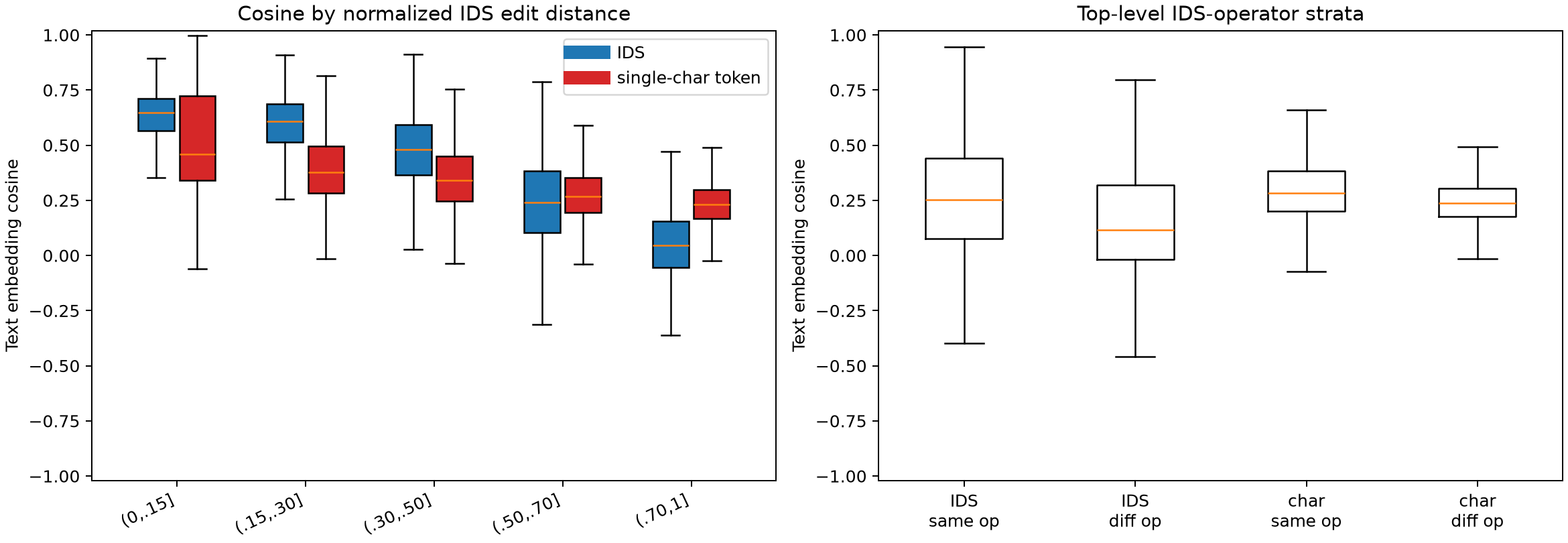}
\caption{\textbf{Text-tower geometry versus symbolic structure (Stage-3 checkpoints).}
\textbf{Left:} text-embedding cosine for $26{,}136$ fixed-seed character pairs grouped by
normalized IDS edit distance (blue: IDS tower; red: single-character tower; box = IQR,
line = median). The IDS tower's cosine falls monotonically as structural distance grows
(medians $0.65\to0.05$; Spearman $\rho=-0.70$), while the single-character tower is
nearly flat ($0.46\to0.23$; $\rho=-0.34$). \textbf{Right:} the same pairs stratified by
shared top-level IDS operator: same vs.\ different medians $0.25$ vs.\ $0.12$ (IDS) and
$0.28$ vs.\ $0.24$ (single-character). Pairs form a distance-stratified diagnostic
sample; distances are flat token edit distances on IDS prefixes.}
\label{fig:geometry}
\end{figure}

\paragraph{Text-tower geometry.}
To test whether the IDS tower organizes its text space by compositional structure, rather
than merely benefiting from a smaller vocabulary, we encode the full $8{,}189$-character
vocabulary with the Stage-3 IDS text tower and with the single-character-token tower of the
otherwise identical checkpoint, and compare embedding geometry against symbolic structure.
Using a fixed-seed, distance-stratified sample of $26{,}136$ character pairs spanning five
bins of normalized IDS token edit distance (flat Levenshtein on the IDS prefix), we correlate pair cosine with structural distance.
The Spearman correlation is $-0.70$ for the IDS tower versus $-0.34$ for the
single-character tower (95\% character-cluster bootstrap CIs $[-0.71,-0.70]$ and
$[-0.35,-0.33]$; difference $-0.36$, CI $[-0.37,-0.35]$; paired permutation
$p<10^{-3}$). Figure~\ref{fig:geometry} (left) shows where the gap comes from: IDS-tower
median cosine falls monotonically from $0.65$ for near-identical decompositions to $0.05$
for unrelated characters, while the single-character tower remains nearly flat
($0.46\to0.23$), so structural proximity organizes only the IDS space. Stratifying by
top-level composition operator (right), same-operator pairs embed markedly closer than
different-operator ones for the IDS tower ($0.25$ vs.\ $0.12$) but not for the
single-character tower ($0.28$ vs.\ $0.24$). Together with the detection results, this
shows that the IDS tower's advantage is accompanied by a structure-aligned representation,
not merely a smaller output space.

\section{Number of Prototypes: $K$ Ablation}
\label{app:k}
The readout clusters each character's normal crops into $K$ prototypes. Table~\ref{tab:ksweep}
sweeps $K$ on the gold test set ($N{=}7{,}358$, $642$ positives) with the frozen $90$th-percentile
readout of the Stage-3 model. F1 peaks at $K{=}10$ ($0.844$) and is flat over $K\in[8,16]$
($0.834$--$0.844$), so the readout is insensitive to $K$: small $K$ (1--2) underfits the
multi-modal legal writings, while large $K$ fragments prototypes and lowers precision. We use
$K{=}10$ throughout.

\begin{table}[h]
\centering
\caption{$K$ sweep on the gold test set ($N{=}7{,}358$, $642$ positives), frozen $90$th-percentile
readout, Stage-3 model.}
\label{tab:ksweep}
\small
\begin{tabular}{rrrrrl}
\toprule
$K$ & $\delta$ & P & R & F1 & TP/FP/FN/TN \\
\midrule
1  & 0.0408 & 0.775 & 0.804 & 0.789 & 516/150/126/6566 \\
2  & 0.0346 & 0.803 & 0.799 & 0.801 & 513/126/129/6590 \\
4  & 0.0284 & 0.836 & 0.815 & 0.825 & 523/103/119/6613 \\
6  & 0.0253 & 0.813 & 0.827 & 0.820 & 531/122/111/6594 \\
8  & 0.0231 & 0.827 & 0.847 & 0.837 & 544/114/98/6602 \\
\textbf{10} & \textbf{0.0214} & \textbf{0.830} & \textbf{0.858} & \textbf{0.844} & \textbf{551/113/91/6603} \\
12 & 0.0201 & 0.812 & 0.857 & 0.834 & 550/127/92/6589 \\
14 & 0.0191 & 0.802 & 0.872 & 0.836 & 560/138/82/6578 \\
16 & 0.0180 & 0.803 & 0.877 & 0.838 & 563/138/79/6578 \\
18 & 0.0172 & 0.794 & 0.883 & 0.836 & 567/147/75/6569 \\
20 & 0.0164 & 0.778 & 0.885 & 0.828 & 568/162/74/6554 \\
\bottomrule
\end{tabular}
\end{table}

\section{Baseline Prompts and Protocols}
\label{app:prompts}
The three general VLMs use one shared prompt with per-character crops; TextPecker \citep{zhu2026textpecker} uses its own
released protocol.

\paragraph{General VLMs (Qwen3.8-Max \citep{qwen2026qwen38max} / GPT-5.6-sol \citep{openai2026gpt56} / Gemini~3.7~Flash \citep{googledeepmind2026gemini37flash}).} All three are called
through their official APIs with the \emph{same} target-conditioned prompt: each crop is presented
together with its \emph{target character} $c$, and the model only judges whether the crop is a
correct rendering of $c$ (yes/no), without having to recognize which character it is. This matches
our verifier's setting (both are conditioned on the target character), keeping the comparison fair.
Prompt text (\texttt{\_\_CHAR\_\_} is replaced by the target character at query time):
\begin{quote}\small\ttfamily\noindent
The target Chinese character for this image crop is: \_\_CHAR\_\_.
The crop may contain parts of neighboring characters, background or banner graphics; ignore
everything except the glyph of the target character itself. Judge whether that glyph is a
CORRECT rendering of ``\_\_CHAR\_\_'': every stroke and the overall structure must match the
standard glyph, with no missing, extra, broken or misplaced strokes. Font style, weight,
color, texture and background are NOT errors, and partial marks of OTHER characters are NOT
errors. Output strictly one JSON object: \{"render\_error": true/false, "reason":
"\textless brief reason\textgreater"\}
\end{quote}
A ``rendering error'' verdict is the positive class. The typical VLM failure is to call a
structurally wrong glyph correct (Sec.~\ref{sec:exp}, Figure~\ref{fig:teaser}).

\paragraph{TextPecker.} TextPecker~\citep{zhu2026textpecker} uses its own released evaluation protocol
(crop-image input, structural anomalies flagged with special markers ``$<\#>$''); we run its two released weights (Qwen3-VL-8B and InternVL3-8B) as
structural-anomaly assessors and refer to the original work for prompt details.

\section{Benchmark Annotation and Additional Qualitative Results}
\label{app:qual}
\paragraph{Annotation protocol.} Annotation proceeds in two steps. A CRAFT \citep{baek2019character} {+} PP-OCRv6 \citep{zhang2026ppocrv6} pipeline
first proposes a bounding box and a character identity for each glyph; annotators then verify the
box, confirm the identity, and label whether the glyph is correctly rendered, counting any stroke-
or component-level deviation as an error while not penalizing font, calligraphic, or artistic
style. Each character thereby receives its (position, identity, correctness) triple.

\begin{figure}[h]
\centering
\includegraphics[width=\linewidth]{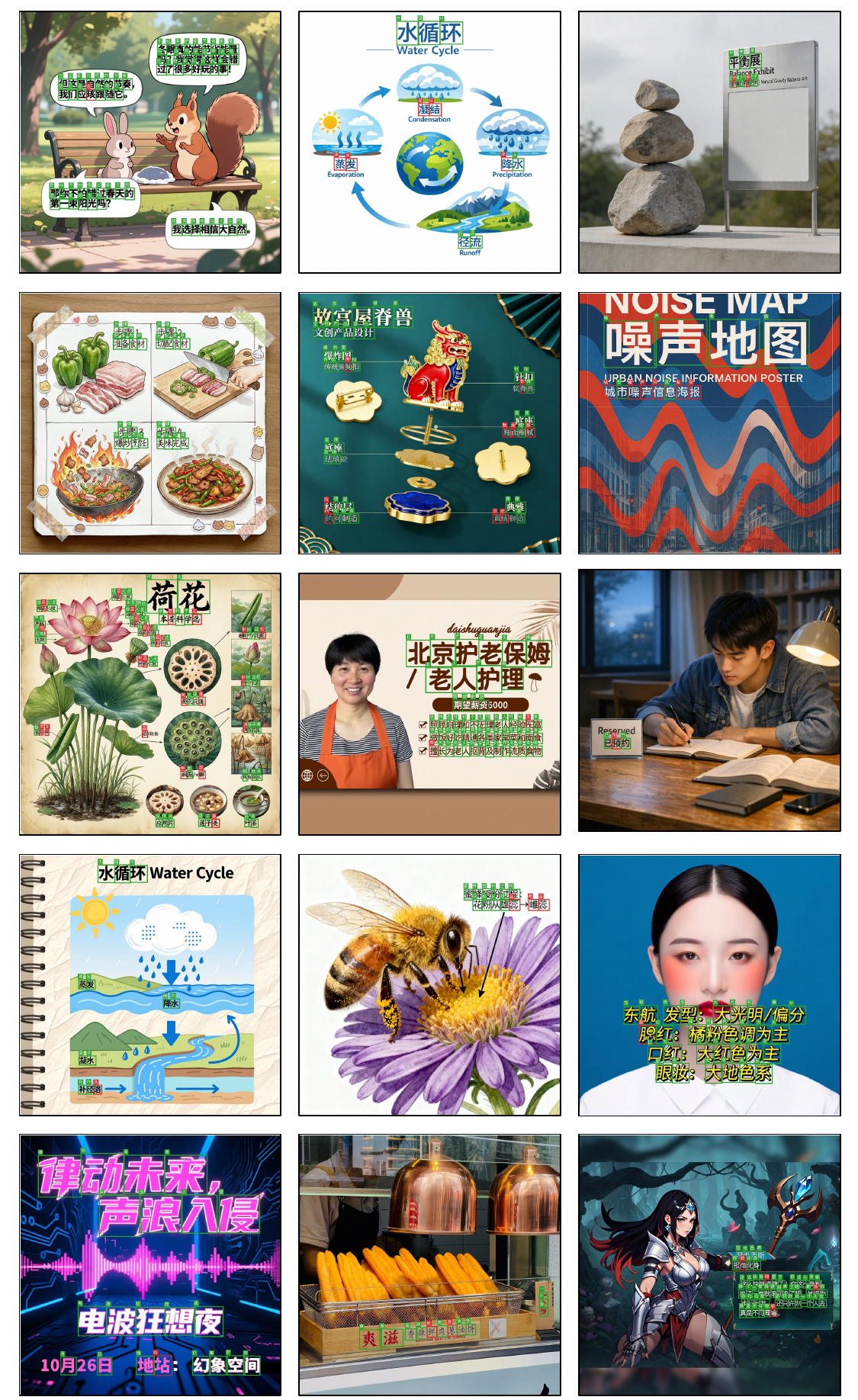}
\caption{Additional \bench{} annotated examples.}
\label{fig:bench_more}
\end{figure}

\paragraph{More benchmark annotations.} Figure~\ref{fig:bench_more} adds \bench{} samples beyond
the Sec.~\ref{sec:exp} example, covering different error types (missing/extra strokes, misplaced
components, wrong characters) and layouts.

\begin{figure}[h]
\centering
\includegraphics[width=\linewidth]{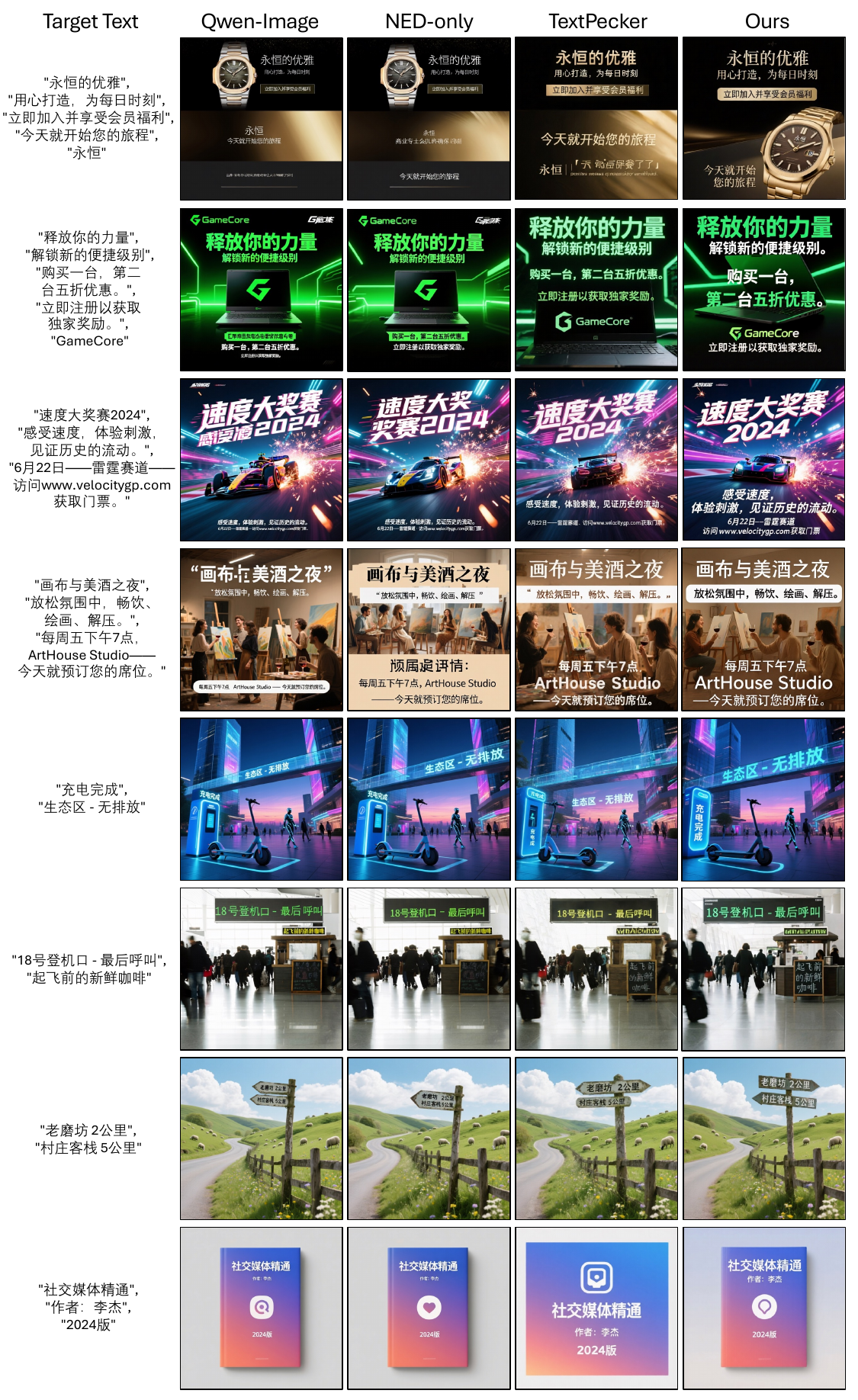}
\caption{Additional Flow-GRPO qualitative comparisons on oneID-ZH \citep{chang2025oneigbench} bench. }
\label{fig:grpo_more}
\end{figure}

\paragraph{More Flow-GRPO comparisons.} Figure~\ref{fig:grpo_more} places Base~(Qwen-Image \citep{wu2025qwenimage}) / NED-only / TextPecker \citep{zhu2026textpecker}
/ Ours side by side for several shared prompts, showing that our reward matches TextPecker \citep{zhu2026textpecker} on glyph
correction while perturbing background, color, and layout less (consistent with the diversity and
aesthetics numbers in Sec.~\ref{sec:rewardexp}).

\end{document}